\documentclass{article}
\usepackage{spconf}

\usepackage{graphicx}
\usepackage{amsmath}
\usepackage{amssymb}
\usepackage{booktabs}
\usepackage{multirow}
\usepackage{subfigure}

\usepackage{color, colortbl}
\usepackage{xcolor}

\usepackage[pagebackref=true,breaklinks=true,letterpaper=true,colorlinks,bookmarks=false]{hyperref}

\usepackage[capitalize]{cleveref}
\usepackage{algorithm}
\usepackage{algpseudocode}

\usepackage{mathtools}
\DeclareMathOperator*{\argmax}{arg\,max}

\title{Robust Feature Learning Against Noisy Labels}
\name{Tsung-Ming Tai $^{1,2}$, Yun-Jie Jhang $^3$, Wen-Jyi Hwang $^4$}
\address{$^1$ Free University of Bozen-Bolzano, $^2$ NVIDIA AI Technology Center \\
$^3$ International Intercollegiate Ph.D. Program, National Tsing Hua University\\
$^4$ Department of Computer Science and Information Engineering, National Taiwan Normal University}

\begin{document}
%\ninept
%
\maketitle
\begin{abstract}
Supervised learning of deep neural networks heavily relies on large-scale datasets annotated by high-quality labels. In contrast, mislabeled samples can significantly degrade the generalization of models and result in memorizing samples, further learning erroneous associations of data contents to incorrect annotations. To this end, this paper proposes an efficient approach to tackle noisy labels by learning robust feature representation based on unsupervised augmentation restoration and cluster regularization. In addition, progressive self-bootstrapping is introduced to minimize the negative impact of supervision from noisy labels. Our proposed design is generic and flexible in applying to existing classification architectures with minimal overheads. Experimental results show that our proposed method can efficiently and effectively enhance model robustness under severely noisy labels.

% Previous works tackled the noisy label problem either by modeling the prior or posterior label transitions or iteratively filtering out the mislabeled samples. 
\end{abstract}
\begin{keywords}
Image classification, noisy labels, robust feature learning.
\end{keywords}

\section{Introduction}
Numerous advancements in deep learning can be attributed to supervised learning on well-labeled datasets. Accurate labeling is critical in deriving informative gradients for model updates targeting optimization. However, noisy labels can significantly bias gradient estimations, leading to incorrect mapping between data content and noisy annotation. Obtaining high-quality annotations is a challenging endeavor. In some applications, the presence of noisy labels is inevitable due to the absence of ground truths. Noisy labels are also commonly found in crowd-sourced or web-crawled datasets \cite{xiao2015learning,kuznetsova2018open}. Recent studies indicate that widely used datasets for image benchmarking also suffer from noisy labels \cite{beyer2020we}.

Earlier works addressed noisy labels by estimating the transitions of noise corruption to recover actual labels \cite{patrini2017making,xia2019anchor}. However, these underlying noise transitions are generally intractable and challenging to estimate accurately. Some previous works explored a surrogate training objective to minimize the impact of noisy labels. For instance, additional training with mean absolute error estimation could have models less sensitive to noise outliers \cite{zhang2018generalized}. Adding a symmetric component to cross-entropy could prevent overlooking noise samples \cite{wang2019symmetric}. More recently, inspired by the lottery ticket hypothesis, researchers found that model parameters could be updated differently based on the gradient sensitivity of training samples \cite{xia2021robust}.

Other prior studies proposed segregating clean and mislabeled samples by either removing them from the training set \cite{chen2019understanding,pleiss2020identifying,han2018co} or by resetting the labels and integrating semi-supervised learning into the training loop \cite{li2020dividemix,kong2019recycling}. However, these methods assumed that samples with smaller loss values likely came from clean labels under the \textit{small loss criterion} \cite{arpit2017closer}. A similar observation was discussed in \cite{xu2019frequency}, where neural networks tend to fit the generic pattern early in training and memorize the hard samples at the end. Despite the evident progress these works have shown over the years, there remains a risk of mistakenly removing clean samples, potentially harming generalizations. Moreover, these methods require more computes by duplicating the classification model \cite{han2018co,li2020dividemix,zhang2020distilling}, or multiple data passes \cite{tanaka2018joint,pleiss2020identifying,han2018co}.

To this end, we propose robust feature learning against noisy labels. Our method introduces two regularization strategies to the training using unsupervised approaches: (i) Augmentation restoration encourages the model to revert the augmented input images, thereby learning to construct robust and rich semantic features encoded by the image backbone; (ii) Clustering regularization groups similar semantic features into sets via differentiable information maximization. Utilizing heuristic evidence that shows neural networks learning generic patterns before memorizing noises, we progressively shift the supervision of the learning target from noisy labels to self-predictions. Our proposed methods do not select training samples, require no extra data passes, and involve minimal additional parameters, distinguishing them from previous works.

\section{Preliminary}
Consider an input image $x$ and corresponding actual labels $y_{clean}$. Let $f_\theta(.)$ denotes an image classification backbone, parameterized by $\theta$, and $g_\phi(.)$ represents a classifier that maps intermediate feature representations to categories using trainable parameters $\phi$. The prediction $\hat{y}$ conditional on the input $x$ can be computed by 
\begin{eqnarray} 
\hat{y} = g_\phi(f_\theta(x)). 
\end{eqnarray}
In practice, To prevent overfitting, we can replace the input $x$ with $T(x)$, where $T(.)$ predefines augmentation transforms.

The objective is to minimize empirical risk by comparing the cross-entropy between $\hat{y}$ and $y_{clean}$. 
\begin{eqnarray} 
L_{task}(\hat{y}, y_{clean}) = - y_{clean} \log \hat{y}, 
\label{eq:clean_task} 
\end{eqnarray} 
where $\hat{y}$ represents the softmax outputs, and $y_{clean}$ denotes the ground-truth labels presented in one-hot encoding.

It is important to note that in the noisy label problem, the ground-truth labels $y_{clean}$ are inaccessible. Instead, only $y_{noisy}$ is observed. Additionally, the actual transition from $y_{clean}$ to $y_{noisy}$ is unknown.

\section{Proposed Method}
\label{sec:method}
We propose a method to enhance feature robustness encoded by the classification backbone to mitigate the negative impact of noisy labels. We encourage the model to reverse the augmented input $T(x)$ back to the original image $x$ using intermediate features by an auxiliary decoder $D_\psi(.)$: 
\begin{eqnarray} 
\hat{x} = D_\psi(f(T(x))) 
\end{eqnarray} 
and minimize the mean square errors between the reconstructed and original image, 
\begin{eqnarray} 
L_{rec}(\hat{x}, x) = || \hat{x} - x ||^2_2. 
\label{eq:rec} 
\end{eqnarray}

To efficiently reverse augmentation transformations on intermediate features, the backbone model must abstract data's structural information, maintaining it within the encoded representations. For instance, \textit{inpainting} is the reversal of \textit{cutout} augmentation, which requires the model to extrapolate the missing information from an image. To ensure sufficient diversity, we adopt RandAugment \cite{cubuk2020randaugment} for $T(.)$. This leads to generating a wide range of proxy tasks on-the-fly during training. It is analogous to a denoising auto-encoder \cite{vincent2008extracting} when viewing the augmentations $T(.)$ as input noise perturbations.

Beyond augmentation restoration, we introduce clustering regularization to further enhance feature robustness. The concept is to group similar features, making the model less sensitive to noise outliers (i.e., mislabeled samples). We employ differentiable \textit{regularized information maximization} (RIM \cite{krause2010discriminative}) to train jointly with the classification task, with an additional cluster head $g_\xi(.)$, which maps the encoded features to the cluster centroid $c = g_\xi(f(x))$. The general form of RIM involves minimizing the following:
\begin{eqnarray} 
R(\theta, \xi) - \lambda I(x; c), 
\end{eqnarray} 
where $I(x; c)$ represents the mutual information between the input $x$ and resulting clusters $c$, and $R(\theta)$ is a regularization penalty for the weights $\theta$ and $\xi$.

The mutual information $I(x; c)$ can be rewritten using the entropy $H(.)$ function: 
\begin{eqnarray} 
I(x; c) = H(c) - H(c|x) 
\end{eqnarray}

Maximizing the abovementioned mutual information is equivalent to (i) constraining the cluster results to a uniform distribution (i.e., by increasing $H(c)$) to avoid the trivial solution, and also (ii) categorizing the input $x$ into different clusters $c$ (i.e., by decreasing $H(c|x)$). We can implement $H(c)$ by minimizing the \textit{Kullback-Leibler} divergence, comparing the clustering result with the uniform distribution as $-KL[c||\mathcal{U}]$. On the other hand, $H(c|x)$ can be directly computed based on the entropy function $H(c|x) = - P(c|x) \log P(c|x)$.

The augmentations can be leveraged for the regularization term $R(\theta, \xi)$. Similar to adversarial training \cite{goodfellow2014explaining}, we ask the model to produce consistent outputs between the original and augmented images. In summary, the cluster loss is defined as:
\begin{eqnarray}
L_{cluster} &=& R(\theta, \xi) - \lambda I(x; c), \label{eq:cluster}\\
            &=& R(\theta, \xi) - \lambda [ H(c) - H(c|x) ], \\
     \nonumber\\
     where~R(\theta, \xi) &=& - g_\xi(f_\theta(x)) \log g_\xi(f_\theta(T(x))), \\
     H(c) &=& -KL[g_\xi(f_\theta(x))||\mathcal{U}], \\
     H(c|x) &=& -P(c|x) \log P(c|x).
\end{eqnarray}

Lastly, we introduce progressive self-bootstrapping, which gradually shifts the supervision target from noisy labels to self-predictions by decaying $\alpha$ from 1 to 0, as follows: 
\begin{eqnarray} 
\hat{y} &=& \log g_\phi(f_\theta(x)),\\
L_{bootstrap} &=& - \alpha~y_{noisy} \log \hat{y} \nonumber \\
&& - (1-\alpha)~(1 \cdot \argmax \hat{y}) \log \hat{y} \label{eq:bootstrap} 
\end{eqnarray}

The overall loss is the sum of each objective. \begin{eqnarray} 
L_{all} = L_{bootstrap} + L_{rec} + L_{cluster} 
\end{eqnarray}
Note that both $L_{rec}$ and $L_{cluster}$ are independent of $y_{noisy}$. By shifting the learning focus from noisy labels to self-predictions, conditional on the learned features, via $L_{bootstrap}$, we can effectively avoid overfitting the mislabeled samples.

\section{Experiment Results}
\begin{table*}
\caption{Experiment results on CIFAR10 under different symmetric and asymmetric noise levels.}
  \setlength{\tabcolsep}{1.8\tabcolsep}
  \label{tab:noisy_cifar10}
  \centering
  \small
  \scalebox{0.97}{
  \begin{tabular}{lc|c|cccc|cccc}
    \toprule
    \multirow{2}{*}{Methods} &
    &
    \multirow{2}{*}{Clean} &
      \multicolumn{4}{c}{Symmetry} &
      \multicolumn{4}{c}{Asymmetry} \\
      & & & 0.2 & 0.4 & 0.6 & 0.8 & 0.1 & 0.2 & 0.3 & 0.4 \\
      \hline
     
    \multirow{2}{*}{CE} & \it{best} & 93.26 & 82.46 & 78.73 & 68.30 & 38.42 & 88.25 & 85.63 & 83.82 & 78.31 \\
                             & \it{last} & 93.08 & 80.88 & 62.56 & 39.76 & 16.94 & 87.87 & 81.56 & 70.19 & 56.87 \\
    % \hline
    % \multirow{2}{*}{CE (RandAugment)} & \it{best} & {\bf 95.74} & 89.46 & 84.88 & 77.38 & 23.49 & 92.06 & 91.21 & 90.25 & 87.55 \\
    %                               & \it{last} & {\bf 95.66} & 85.25 & 67.31 & 45.73 & 18.60 & 90.90 & 83.48 & 72.27 & 61.78 \\
    \hline
    \multirow{2}{*}{Forward \cite{patrini2017making}} & \it{best} & \textbf{95.64} & 92.39 & 87.87 & 77.21 & 22.38 & 92.22 & 85.87 & 65.53 & 17.51 \\
                             & \it{last} & 95.30 & 92.35 & 81.19 & 45.71 & 10.02 & 90.30 & 79.12 & 64.47 & 8.48 \\
    \hline
    \multirow{2}{*}{GCE \cite{zhang2018generalized}} & \it{best} & 92.35 & 90.82 & 88.24 & 83.20 & 59.50 & 91.51 & 90.62 & 89.94 & 88.85 \\
                                              & \it{last} & 92.19 & 90.81 & 88.02 & 82.58 & 58.96 & 91.37 & 90.48 & 89.71 & 88.71 \\
    \hline
    \multirow{2}{*}{Bootstrap \cite{reed2014training}} & \it{best} & 95.63 & 86.54 & 80.00 & 74.22 & 51.47 & 91.09 & 87.02 & 85.88 & 82.36 \\
                              & \it{last} & \textbf{95.51} & 86.54 & 69.39 & 45.98 & 21.03 & 90.66 & 83.35 & 71.76 & 60.59 \\
    \hline
    \multirow{2}{*}{IMAE \cite{wang2019imae}} & \it{best} & 91.37 & 89.93 & 86.25 & 76.71 & 36.62 & 90.35 & 88.66 & 85.96 & 82.36 \\
                          & \it{last} & 91.35 & 88.87 & 85.87 & 68.00 & 31.20 & 90.20 & 88.22 & 82.72 & 60.59 \\
    \hline
    \multirow{2}{*}{SL \cite{wang2019symmetric}} & \it{best} & 95.33 & 91.09 & 87.11 & 80.13 & 44.90 & 91.78 & 89.73 & 87.09 & 77.45 \\
                                            & \it{last} & 95.24 & 90.95 & 86.06 & 65.76 & 24.27 & 91.51 & 88.69 & 74.83 & 60.52 \\
    \hline
    \multirow{2}{*}{JO \cite{tanaka2018joint}} & \it{best} & 95.26 & 90.01 & 85.90 & 78.41 & 59.41 & 92.08 & 91.27 & 90.41 & 87.56 \\
                                        & \it{last} & 95.17 & 87.56 & 76.95 & 56.87 & 28.35 & 91.46 & 84.17 & 74.43 & 62.50 \\
    \hline
    \multirow{2}{*}{Ours} & \it{best} & 95.37 & {\bf 94.42} & {\bf 93.02} & {\bf 90.75} & {\bf 85.83} & {\bf 95.17} & {\bf 95.05} & {\bf 94.74} & {\bf 94.64} \\
                          & \it{last} & 95.12 & {\bf 94.34} & {\bf 92.72} & {\bf 90.52} & {\bf 85.22} & {\bf 94.95} & {\bf 94.74} & {\bf 94.52} & {\bf 94.49} \\
    \bottomrule
  \end{tabular}
    }
\end{table*}
\begin{table}
  \caption{Experiment results on CIFAR100 dataset.}
  \label{tab:noisy_cifar100}
  \centering
  \small
  \scalebox{0.98}{
  \begin{tabular}{l|c|cccc}
    \toprule
    \multirow{2}{*}{Methods} &
    \multirow{2}{*}{Clean} &
      \multicolumn{4}{c}{Symmetry} \\
      & & 0.2 & 0.4 & 0.6 & 0.8 \\
      \midrule
    CE & \textbf{75.41} & 58.72 & 48.20 & 37.41 & 18.10 \\ 
    Forward \cite{reed2014training} & - & 63.16 & 54.65 & 44.62 & 24.83 \\
    GCE \cite{zhang2018generalized} & - & 67.61 & 62.64 & 53.16 & 29.16 \\
    PENCIL \cite{yi2019probabilistic} & - & \textbf{73.86} & 69.12 & 57.79 & fail \\
    \midrule
    Ours & 75.34 & 72.20 & \textbf{69.17} & \textbf{62.98} & \textbf{40.09} \\
    \bottomrule
  \end{tabular}
  }
\end{table}
\begin{table*}
  \caption{Experiment results on Clothing1M dataset.}
  \centering
  \small
  \scalebox{0.98}{
  \begin{tabular}{c|ccccccccc}
    \toprule
     & CE & Forward \cite{patrini2017making} & GCE \cite{zhang2018generalized} & Bootstrap \cite{reed2014training} & IMAE \cite{wang2019imae} & SL \cite{wang2019symmetric} & JO \cite{tanaka2018joint} & PENCIL \cite{yi2019probabilistic} & Ours \\ 
    \midrule
    Acc. (\%) & 68.80 & 69.84 & 69.75 & 68.94 & 73.20 & 71.02 & 72.23 & 73.49 & {\bf 73.67} \\
    \bottomrule
  \end{tabular}
  }
  \label{tab:clothing}
\end{table*}
\begin{table}[h]
    \centering
    \small
    \caption{Analysis on each individual proposed design.}
    \scalebox{0.98}{
    \begin{tabular}{lcc}
    \toprule
    Component & Symmetric 0.6 & Symmetric 0.8 \\
    \midrule
    CE & 68.30 & 38.42\\
    % CE (RandAugment) & 77.38 & 23.49\\
    \midrule
    + (A) $L_{bootstrap}$ (eq~(\ref{eq:bootstrap})) & 88.82 & 16.06\\
    + (B) $L_{rec}$ (eq~(\ref{eq:rec})) & 79.63 & 58.92\\
    + (C) $L_{cluster}$ (eq~(\ref{eq:cluster})) & 78.91 & 21.91\\
    \midrule
    + (A) + (B) & 88.57 & 45.30\\
    + (A) + (C) & 89.73 & 19.51\\
    + (A) + (B) + (C) & {\bf 90.75} & {\bf 85.83}\\
    \bottomrule
    \end{tabular}
    }
    \label{tab:ablation1}
\end{table}

We adopt noise model definitions from \cite{chen2019understanding}. For $C$-class classification, {\it symmetric noise} relabels $\epsilon$ of the labels in each category and uniformly distributed among the remaining $C-1$ classes. In contrast, {asymmetric noise} replaces $\epsilon$ of the ground-truth labels in each category with the class id of the subsequent class (i.e., class 0 becomes class 1, and so on).

All experiments use ResNet18 \cite{he2016deep} as the classification backbone. Training epochs are set to 300 for CIFAR10 and CIFAR100, and 50 for Clothing1M \cite{xiao2015learning}. We employ the SGD optimizer with momentum of 0.9, learning rate of 0.1, and weight decay of 1e-4. We decrease the learning rate at 50\% and 75\% of training progress by step ratio of 0.1. We implemented our model in PyTorch and conducted experiments on a single RTX 3090 GPU. The batch size is 128 for CIFAR10 and CIFAR100, and 32 for Clothing1M. To ensure meaningful comparisons, we compare our results with methods that do not require additional data passes or co-training architecture. All the reported numbers in the comparisons are either directly obtained from their respective source papers, if available, or carefully reproduced based on their released implementations.

\begin{figure*}
\centering
\includegraphics[width=.28\linewidth]{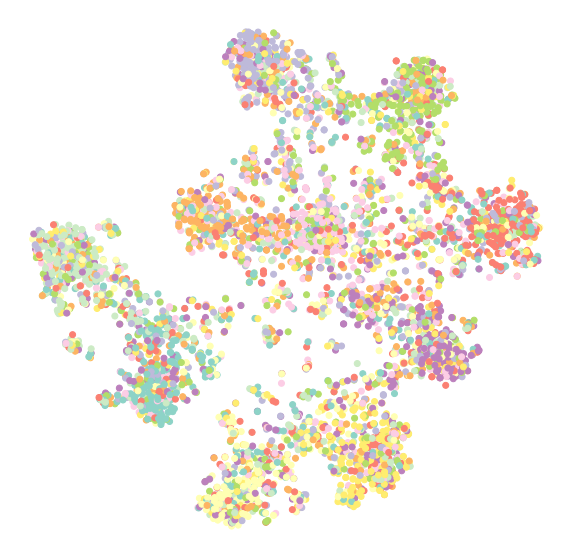}
\includegraphics[width=.28\linewidth]{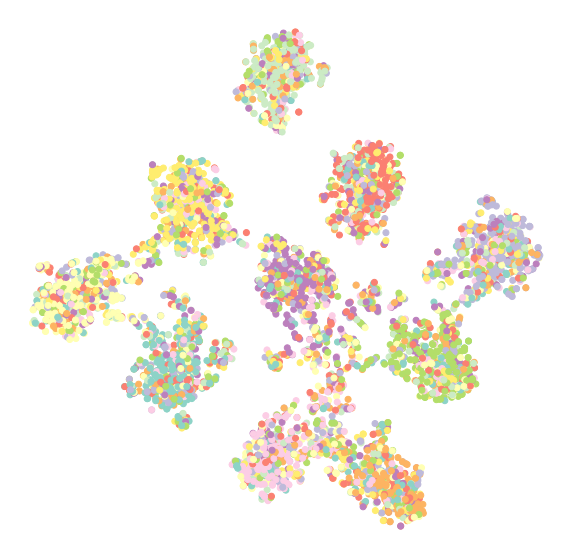}
\includegraphics[width=.28\linewidth]{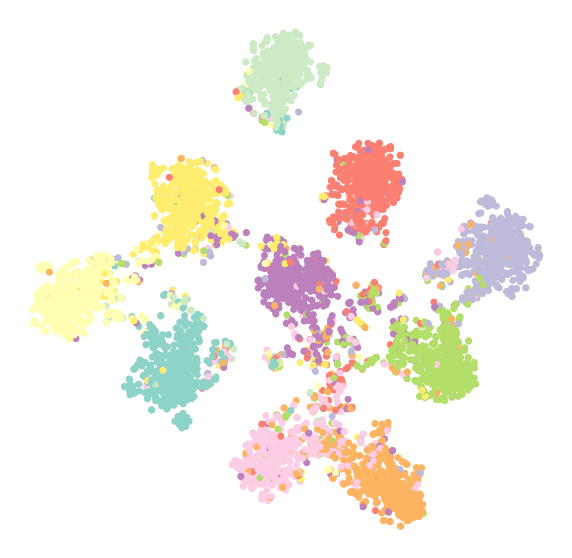}
   \caption{The t-SNE visualization of encoded features on CIFAR10 (symmetric noise, $\epsilon$=0.6). \textit{Left:} Proposed method without $L_{cluster}$. \textit{Middle:} Proposed method and colored by noisy labels. \textit{Right:} Proposed method and colored by actual labels.}
\label{fig:imsat}
\end{figure*}

\begin{figure}
\centering
\includegraphics[width=.48\linewidth]{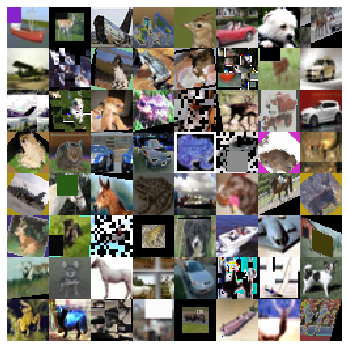}
\includegraphics[width=.48\linewidth]{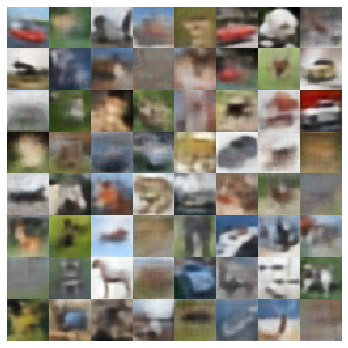}
   \caption{Demonstration of augmentation restoration. \textit{Left:} Augmented images. \textit{Right:} Predicted restorations.}
\label{fig:self_deaugment}
\end{figure}

\subsection{CIFAR10 Results} 
Table~\ref{tab:noisy_cifar10} compares our proposed method with various baselines under symmetric and asymmetric noise ratios on CIFAR10. We report both the best and the last epoch validation accuracy. It is important to note that the best validation accuracy is only considered when noise-free annotation is available to determine the optimal model checkpoint. The {\it clean} configuration is included as a benchmark to measure the performance while no noisy labels are presented. Our method demonstrates robust performance across diverse noise levels, particularly under severe conditions such as a symmetric noise level of $\epsilon$=0.8 and an asymmetric noise level of $\epsilon=0.4$, while maintaining the same accuracy level on clean labels.

\subsection{CIFAR100 Results} 
Table~\ref{tab:noisy_cifar100} depicts the results of our method applied to CIFAR100. Notably, the PENCIL method results in divergence and performs no better than a random guess under severe symmetric noise conditions, specifically $\epsilon=0.8$, where 80\% of the samples are mislabeled and distributed among the other 99 classes. Despite this, our method can still learn meaningful information under severely noisy labels.

\subsection{Clothing1M Results} 
Table~\ref{tab:clothing} evaluates our method on Clothing1M, a web-crawler labeled dataset. The results indicate that our approach can be effectively applied to practical noisy label scenarios and outperform the baseline without requiring further hyperparameter tuning. Note that Clothing1M also suffers from label imbalance. We mark this as a potential future work in the noisy label problem.

\subsection{Ablation Study} 
We conducted an ablation study on CIFAR10 with symmetric noise ratios of $\epsilon=0.6$ and $\epsilon=0.8$. Table \ref{tab:ablation1} shows the individual contributions of each design aspect in our proposed method. Firstly, we initiated the process with a vanilla cross-entropy baseline. Subsequently, incorporating either progressive self-bootstrapping $L_{bootstrap}$, augmentation restoration $L_{rec}$, or cluster regularization $L_{cluster}$ improved robustness under a noise ratio of $\epsilon=0.6$. Nevertheless, only augmentation restoration $L_{rec}$ improved the more severe noise ratio of $\epsilon=0.8$. This suggests that the proposed augmentation restoration effectively enhances training robustness. Lastly, the optimal configuration is attained by combining all the proposed components, yielding an absolute accuracy improvement of 12.78\% and 47.41\% on $\epsilon=0.6$ and $\epsilon=0.8$, respectively, compared to the cross-entropy baseline.

Figure \ref{fig:imsat} visualizes the encoded features of our method using t-SNE \cite{maaten2008visualizing}. The leftmost figure displays the features trained without $L_{cluster}$. The middle figure shows the inclusion of $L_{cluster}$ in training can divide the samples into groups. Actual labels then color the rightmost figure. Upon comparing the middle and rightmost figures, it is evident that our method corrected mislabeled samples and achieved consistency with the cluster criterion. Indeed, unsupervised clustering aids in the formation of robust features without interaction with noisy labels.

Figure \ref{fig:self_deaugment} presents the augmented images alongside their corresponding restoration predictions. While some augmentations pose a challenge to recover due to the corruption of critical features, it is evident that the model learned structural information. For instance, the top-left ship image successfully recovers the missing square. Instances of color adjustments and noise removals can also be identified in the figure. This evidence underscores the ability of the proposed augmentation restoration to learn robust representations in an unsupervised manner efficiently.

\section{Conclusions}
This paper introduced an approach for robust feature learning against noisy labels based on unsupervised augmentation restoration and cluster regularization. In addition, progressive self-bootstrapping is utilized to gradually shift the training objective from noisy labels to self-predictions. Combining these designs shows strong performance under various noisy label levels while retaining accuracy with no mislabeled samples.
% This paper introduced an approach for robust feature learning against noisy labels based on unsupervised augmentation restoration and cluster regularization. Furthermore, we employ progressive self-bootstrapping to shift the training objective from noisy labels to self-predictions gradually. Integrating these elements demonstrates strong performance across varying levels of label noise while maintaining accuracy in scenarios devoid of mislabeled samples.

% References should be produced using the bibtex program from suitable
% BiBTeX files (here: strings, refs, manuals). The IEEEbib.bst bibliography
% style file from IEEE produces unsorted bibliography list.
% -------------------------------------------------------------------------
\bibliographystyle{IEEEbib}
\bibliography{refs}

\begin{thebibliography}{10}\itemsep=-1pt

\bibitem{arpit2017closer}
Devansh Arpit, Stanislaw Jastrzebski, Nicolas Ballas, David Krueger, Emmanuel
  Bengio, Maxinder~S Kanwal, Tegan Maharaj, Asja Fischer, Aaron Courville,
  Yoshua Bengio, et~al.
\newblock A closer look at memorization in deep networks.
\newblock In {\em ICML}, pages 233--242. PMLR, 2017.

\bibitem{beyer2020we}
Lucas Beyer, Olivier~J H{\'e}naff, Alexander Kolesnikov, Xiaohua Zhai, and
  A{\"a}ron van~den Oord.
\newblock Are we done with imagenet?
\newblock {\em arXiv:2006.07159}, 2020.

\bibitem{chen2019understanding}
Pengfei Chen, Benben Liao, Guangyong Chen, and Shengyu Zhang.
\newblock Understanding and utilizing deep neural networks trained with noisy
  labels.
\newblock {\em arXiv:1905.05040}, 2019.

\bibitem{cubuk2020randaugment}
Ekin~D Cubuk, Barret Zoph, Jonathon Shlens, and Quoc~V Le.
\newblock Randaugment: Practical automated data augmentation with a reduced
  search space.
\newblock In {\em CVPR Workshops}, pages 702--703, 2020.

\bibitem{goodfellow2014explaining}
Ian~J Goodfellow, Jonathon Shlens, and Christian Szegedy.
\newblock Explaining and harnessing adversarial examples.
\newblock {\em arXiv:1412.6572}, 2014.

\bibitem{han2018co}
Bo Han, Quanming Yao, Xingrui Yu, Gang Niu, Miao Xu, Weihua Hu, Ivor Tsang, and
  Masashi Sugiyama.
\newblock Co-teaching: Robust training of deep neural networks with extremely
  noisy labels.
\newblock In {\em Neurips}, pages 8527--8537, 2018.

\bibitem{he2016deep}
Kaiming He, Xiangyu Zhang, Shaoqing Ren, and Jian Sun.
\newblock Deep residual learning for image recognition.
\newblock In {\em CVPR}, pages 770--778, 2016.

\bibitem{kong2019recycling}
Kyeongbo Kong, Junggi Lee, Youngchul Kwak, Minsung Kang, Seong~Gyun Kim, and
  Woo-Jin Song.
\newblock Recycling: Semi-supervised learning with noisy labels in deep neural
  networks.
\newblock {\em IEEE Access}, 7:66998--67005, 2019.

\bibitem{krause2010discriminative}
Andreas Krause, Pietro Perona, and Ryan Gomes.
\newblock Discriminative clustering by regularized information maximization.
\newblock {\em Neurips}, 23, 2010.

\bibitem{kuznetsova2018open}
Alina Kuznetsova, Hassan Rom, Neil Alldrin, Jasper Uijlings, Ivan Krasin, Jordi
  Pont-Tuset, Shahab Kamali, Stefan Popov, Matteo Malloci, Tom Duerig, et~al.
\newblock The open images dataset v4: Unified image classification, object
  detection, and visual relationship detection at scale.
\newblock {\em arXiv:1811.00982}, 2018.

\bibitem{li2020dividemix}
Junnan Li, Richard Socher, and Steven~CH Hoi.
\newblock Dividemix: Learning with noisy labels as semi-supervised learning.
\newblock {\em arXiv:2002.07394}, 2020.

\bibitem{maaten2008visualizing}
Laurens van~der Maaten and Geoffrey Hinton.
\newblock Visualizing data using t-sne.
\newblock {\em Journal of machine learning research}, 9(Nov):2579--2605, 2008.

\bibitem{patrini2017making}
Giorgio Patrini, Alessandro Rozza, Aditya Krishna~Menon, Richard Nock, and
  Lizhen Qu.
\newblock Making deep neural networks robust to label noise: A loss correction
  approach.
\newblock In {\em CVPR}, pages 1944--1952, 2017.

\bibitem{pleiss2020identifying}
Geoff Pleiss, Tianyi Zhang, Ethan~R Elenberg, and Kilian~Q Weinberger.
\newblock Identifying mislabeled data using the area under the margin ranking.
\newblock {\em arXiv:2001.10528}, 2020.

\bibitem{reed2014training}
Scott Reed, Honglak Lee, Dragomir Anguelov, Christian Szegedy, Dumitru Erhan,
  and Andrew Rabinovich.
\newblock Training deep neural networks on noisy labels with bootstrapping.
\newblock {\em arXiv:1412.6596}, 2014.

\bibitem{tanaka2018joint}
Daiki Tanaka, Daiki Ikami, Toshihiko Yamasaki, and Kiyoharu Aizawa.
\newblock Joint optimization framework for learning with noisy labels.
\newblock In {\em CVPR}, pages 5552--5560, 2018.

\bibitem{vincent2008extracting}
Pascal Vincent, Hugo Larochelle, Yoshua Bengio, and Pierre-Antoine Manzagol.
\newblock Extracting and composing robust features with denoising autoencoders.
\newblock In {\em ICML}, pages 1096--1103, 2008.

\bibitem{wang2019imae}
Xinshao Wang, Yang Hua, Elyor Kodirov, and Neil~M Robertson.
\newblock Imae for noise-robust learning: Mean absolute error does not treat
  examples equally and gradient magnitude's variance matters.
\newblock {\em arXiv:1903.12141}, 2019.

\bibitem{wang2019symmetric}
Yisen Wang, Xingjun Ma, Zaiyi Chen, Yuan Luo, Jinfeng Yi, and James Bailey.
\newblock Symmetric cross entropy for robust learning with noisy labels.
\newblock In {\em ICCV}, pages 322--330, 2019.

\bibitem{xia2021robust}
Xiaobo Xia, Tongliang Liu, Bo Han, Chen Gong, Nannan Wang, Zongyuan Ge, and Yi
  Chang.
\newblock Robust early-learning: Hindering the memorization of noisy labels.
\newblock In {\em ICLR}, 2021.

\bibitem{xia2019anchor}
Xiaobo Xia, Tongliang Liu, Nannan Wang, Bo Han, Chen Gong, Gang Niu, and
  Masashi Sugiyama.
\newblock Are anchor points really indispensable in label-noise learning?
\newblock In {\em Neurips}, pages 6838--6849, 2019.

\bibitem{xiao2015learning}
Tong Xiao, Tian Xia, Yi Yang, Chang Huang, and Xiaogang Wang.
\newblock Learning from massive noisy labeled data for image classification.
\newblock In {\em CVPR}, pages 2691--2699, 2015.

\bibitem{xu2019frequency}
Zhi-Qin~John Xu, Yaoyu Zhang, Tao Luo, Yanyang Xiao, and Zheng Ma.
\newblock Frequency principle: Fourier analysis sheds light on deep neural
  networks.
\newblock {\em arXiv:1901.06523}, 2019.

\bibitem{yi2019probabilistic}
Kun Yi and Jianxin Wu.
\newblock Probabilistic end-to-end noise correction for learning with noisy
  labels.
\newblock In {\em CVPR}, pages 7017--7025, 2019.

\bibitem{zhang2018generalized}
Zhilu Zhang and Mert Sabuncu.
\newblock Generalized cross entropy loss for training deep neural networks with
  noisy labels.
\newblock In {\em Neurips}, pages 8778--8788, 2018.

\bibitem{zhang2020distilling}
Zizhao Zhang, Han Zhang, Sercan~O Arik, Honglak Lee, and Tomas Pfister.
\newblock Distilling effective supervision from severe label noise.
\newblock In {\em CVPR}, pages 9294--9303, 2020.

\end{thebibliography}

\end{document}